\documentclass[11pt,a4paper]{article}
\usepackage[hyperref]{emnlp2018}
\usepackage{times}
\usepackage{latexsym}

\usepackage{url}
\usepackage{color}
\usepackage{amsmath}
\usepackage{graphicx}
\graphicspath{{fig/}}
\usepackage{multirow}

\aclfinalcopy 

\title{
Spider: A Large-Scale Human-Labeled Dataset for Complex and Cross-Domain Semantic Parsing and Text-to-SQL Task
}

\author{Tao Yu \quad\quad Rui Zhang \quad\quad Kai Yang \quad\quad Michihiro Yasunaga \\{\bf Dongxu Wang \quad Zifan Li \quad James Ma \quad Irene Li} \\{\bf \quad Qingning Yao \quad Shanelle Roman \quad Zilin Zhang \quad Dragomir R. Radev}\\
Department of Computer Science, Yale University\\
\scalebox{0.87}[0.9]{{\tt \{tao.yu,\,r.zhang,\,k.yang,\,michihiro.yasunaga,\,dragomir.radev\}@yale.edu}}}

\date{}

\begin{document}
\maketitle
\begin{abstract}
We present \textit{Spider}, a large-scale, complex and cross-domain semantic parsing and text-to-SQL dataset annotated by 11 college students.
It consists of 10,181 questions and 5,693 unique complex SQL queries on 200 databases with multiple tables, covering 138 different domains.
We define a new complex and cross-domain semantic parsing and text-to-SQL task where different complex SQL queries and databases appear in train and test sets.
In this way, the task requires the model to generalize well to both new SQL queries and new database schemas.
Spider is distinct from most of the previous semantic parsing tasks because they all use a single database and the exact same programs in the train set and the test set.
We experiment with various state-of-the-art models and the best model achieves only 12.4\% exact matching accuracy on a database split setting.
This shows that Spider presents a strong challenge for future research.
Our dataset and task are publicly available at \url{https://yale-lily.github.io/spider}.
\end{abstract}

\section{Introduction}
Semantic parsing (SP) is one of the most important tasks in natural language processing (NLP).
It requires both understanding the meaning of natural language sentences and mapping them to meaningful executable queries such as logical forms, SQL queries, and Python code.

Recently, some state-of-the-art methods with Seq2Seq architectures are able to achieve over 80\% exact matching accuracy even on some complex benchmarks such as ATIS and GeoQuery. These models seem to have already solved most problems in this field. 

\begin{figure}[!t]
    \vspace{-1.5mm}\hspace{-1mm}
    \centering
    \includegraphics[width=0.48\textwidth]{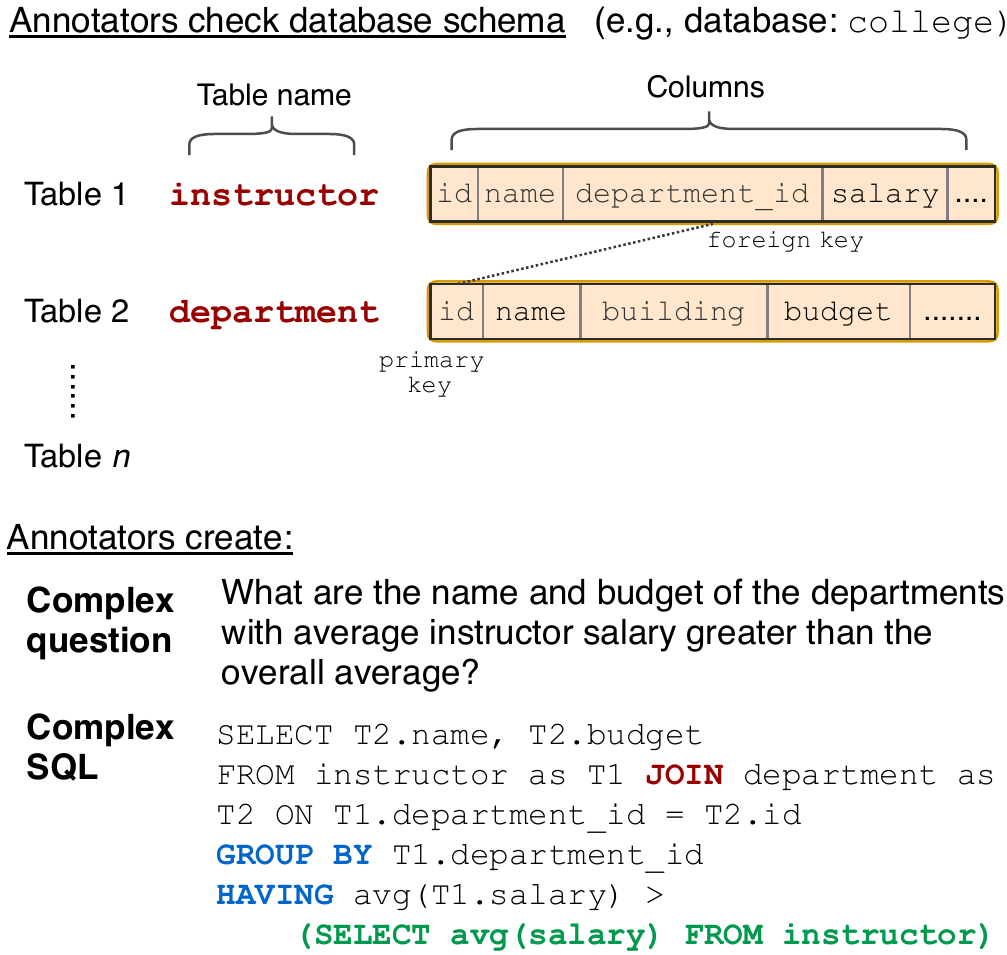}\vspace{-2mm}
    \caption{
    Our corpus annotates complex questions and SQLs. The example contains joining of multiple tables, a \texttt{GROUP BY} component, and a nested query.
    }
\label{fig:task}
\vspace{-6mm}
\end{figure}

However, previous tasks in this field have a simple but problematic task definition because most of these results are predicted by semantic ``matching" rather than semantic parsing. Existing datasets for SP have two shortcomings.
First, those that have complex programs \cite{zelle96,li2014constructing,yaghmazadeh2017sqlizer,iyer17} are too small in terms of the number of programs for training modern data-intensive models and have only a single dataset, meaning that the same database is used for both training and testing the model. More importantly, the number of logic forms or SQL labels is small and each program has about 4-10 paraphrases of natural language problem to expand the size of the dataset. Therefore, the exact same target programs appear in both the train and test sets. The models can achieve decent performances even on very complex programs by memorizing the patterns of question and program pairs during training and decoding the programs exactly the same way as it saw in the training set during testing. \citet{cathy18} split the dataset by programs so that no two identical programs would be in both the train and test sets. They show that the models built on this question-splitting data setting fail to generalize to unseen programs.
Second, existing datasets that are large in terms of the number of programs and databases such as WikiSQL \cite{Zhong2017} contain only simple SQL queries and single tables. In order to test a model's real semantic parsing performance on unseen complex programs and its ability to generalize to new domains, an SP dataset that includes a large amount of complex programs and databases with multiple tables is a must.

However, compared to other large, realistic datasets such as ImageNet for object recognition \cite{imagenet_cvpr09} and SQuAD for reading comprehension \cite{Pranav16}, creating such SP dataset is even more time-consuming and challenging in some aspects due to the following reasons.
First, it is hard to find many databases with multiple tables online. Second, given a database, annotators have to understand the complex database schema to create a set of questions such that their corresponding SQL queries cover all SQL patterns. Moreover, it is even more challenging to write different complex SQL queries. Additionally, reviewing and quality-checking of question and SQL pairs takes a significant amount of time. All of these processes require very specific knowledge in databases. 

To address the need for a large and high-quality dataset for a new complex and cross-domain semantic parsing task, we introduce {\it Spider}, which consists of 200 databases with multiple tables, 10,181 questions, and 5,693 corresponding complex SQL queries, all written by 11 college students spending a total of 1,000 man-hours.
As Figure \ref{fig:task} illustrates, given a database with multiple tables including foreign keys, our corpus creates and annotates complex questions and SQL queries including different SQL clauses such as joining and nested query.
In order to generate the SQL query given the input question, models need to understand both the natural language question and relationships between tables and columns in the database schema.

In addition, we also propose a new task for the text-to-SQL problem. Since Spider contains 200 databases with foreign keys, we can split the dataset with complex SQL queries in a way that no database overlaps in train and test, which overcomes the two shortcomings of prior datasets, and defines a new semantic parsing task in which the model needs to generalize not only to new programs but also to new databases. Models have to take questions and database schemas as inputs and predict unseen queries on new databases. 

To assess the task difficulty, we experiment with several state-of-the-art semantic parsing models. All of them struggle with this task.
The best model achieves only 12.4\% exact matching accuracy in the database split setting.
This suggests that there is a large room for improvement.

\section{Related Work and Existing Datasets}
\label{sec:rel}

Several semantic parsing datasets with different queries have been created. The output can be in many formats, e.g., logic forms. These datasets include ATIS \cite{Price90,Dahl94}, GeoQuery \cite{zelle96}, and JOBS \cite{tang01}. They have been studied extensively \cite{zelle96,Zettlemoyer05,wong07,Das10,Liang11,banarescu13,artzi13,Reddy14,Berant14,dong16}. However, they are domain specific and there is no standard label guidance for multiple SQL queries.

Recently, more semantic parsing datasets using SQL as programs have been created. \citet{iyer17} and \citet{Popescu03} labeled SQL queries for ATIS and GeoQuery datasets. Other existing text-to-SQL datasets also include Restaurants \cite{tang2001using,Popescu03}, Scholar \cite{iyer17}, Academic \cite{li2014constructing}, Yelp and IMDB \cite{Yaghmazadeh17}, Advising \cite{cathy18}, and WikiSQL \cite{Zhong2017}. These datasets have been studied for decades in both the NLP community \cite{warren1982efficient,popescu2003towards,popescu2004modern,li2006constructing,giordani2012translating,wang2017synthesizing,iyer17,Zhong2017,Xu2017,Yu18,pshuang2018PT-MAML,2018executionguided,P18-1068,mccann2018natural} and the Database community \cite{li2014constructing, Yaghmazadeh17}. We provide detailed statistics on these datasets in Table \ref{tb:data}. 

Most of the previous work train their models without schemas as inputs because they use a single database for both training and testing. Thus, they do not need to generalize to new domains. Most importantly, these datasets have a limited number of labeled logic forms or SQL queries. In order to expand the size of these datasets and apply neural network approaches, each logic form or SQL query has about 4-10 paraphrases for the natural language input. Most previous studies follow the standard question-based train and test split \cite{Zettlemoyer05}. This way, the exact same target queries (with similar paraphrases) in the test appear in training set as well. Utilizing this assumption, existing models can achieve decent performances even on complex programs by memorizing database-specific SQL templates. However, this accuracy is artificially inflated because the model merely needs to decide which template to use during testing. \citet{cathy18} show that template-based approaches can get even higher results. To avoid getting this inflated result, \citet{cathy18} propose a new, program-based splitting evaluation, where the exact same queries do not appear in both training and testing. They show that under this framework, the performance of all the current state-of-the-art semantic parsing systems drops dramatically even on the same database, indicating that these models fail to generalize to unseen queries. This indicates that current studies in semantic parsing have limitations.

We also want the model to generalize not only to unseen queries but also to unseen databases.
\citet{Zhong2017} published the WikiSQL dataset. In their problem definition, the databases in the test set do not appear in the train or development sets. Also, the task needs to take different table schemas as inputs. Therefore, the model has to generalize to new databases.
However, in order to generate 80654 questions and SQL pairs for 24241 databases, \citet{Zhong2017} made simplified assumptions about the SQL queries and databases. Their SQL labels only cover single \texttt{SELECT} column and aggregation, and \texttt{WHERE} conditions. Moreover, all the databases only contain single tables. No \texttt{JOIN}, \texttt{GROUP BY}, and \texttt{ORDER BY}, etc. are included.

Recently, researchers have constructed some datasets for code generation including IFTTT \cite{quirk2015language}, DJANGO \cite{Oda15}, HEARTHSTONE \cite{ling16}, NL2Bash \cite{nl2bash}, and CoNaLa \cite{yin18msr}. These tasks parse natural language descriptions into a more general-purpose programming language such as Python \cite{Allamanis15,ling16, RabinovichSK17,Yin17}.

\section{Corpus Construction}
\label{sec:data_collection}
\begin{figure}[t!]
\centering
\includegraphics[width=0.45\textwidth]{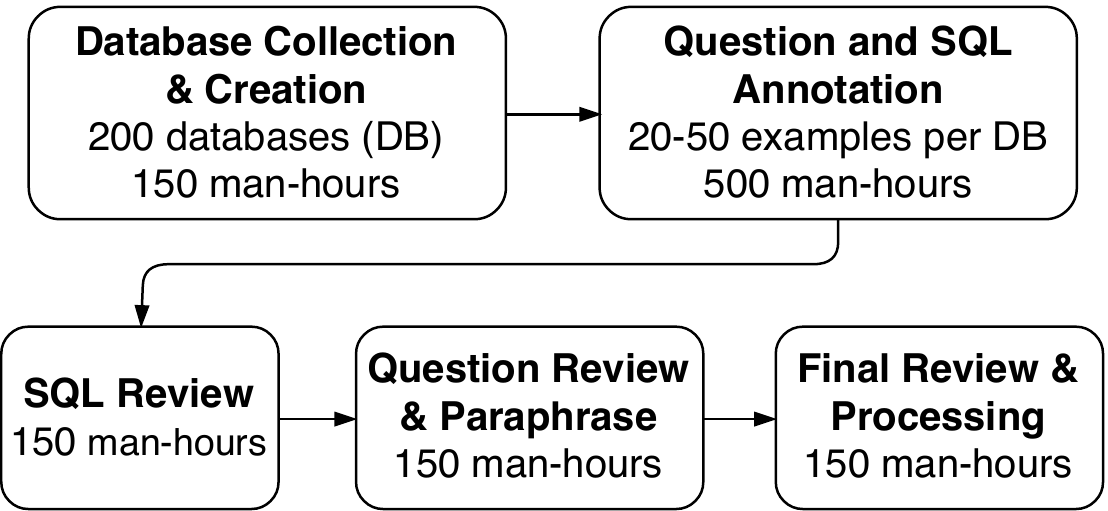}\vspace{-2mm}
\caption{The annotation process of our Spider corpus.}
\label{fig:annotation}\vspace{-3mm}
\end{figure}
All questions and SQL queries were written and reviewed by 11 computer science students. Some of them were native English speakers.
As illustrated in Figure \ref{fig:annotation}, we develop our dataset in five steps, spending around 1,000 hours of human labor in total: \S \ref{sec:database_collection} Database Collection and Creation, \S \ref{sec:question_and_sql_annnotation} Question and SQL Annotation, \S \ref{sec:sql_review} SQL Review, \S \ref{sec:question_review} Question Review and Paraphrase, \S \ref{sec:final_question} Final Question and SQL Review.

\subsection{Database Collection and Creation}
\label{sec:database_collection}
Collecting databases with complex schemas is hard. Although relational databases are widely used in industry and academia, most of them are not publicly available. Only a few databases with multiple tables are easily accessible online.

Our 200 databases covering 138 different domains are collected from three resources. First, we collected about 70 complex databases from different college database courses, SQL tutorial websites, online csv files, and textbook examples. Second, we collected about 40 databases from the DatabaseAnswers\footnote{\url{http://www.databaseanswers.org/}} where contains over 1,000 data models across different domains. These data models contain only database schemas. We converted them into SQLite, populated them using an online database population tool\footnote{\url{http://filldb.info/}}, and then manually corrected some important fields so that the table contents looked natural. Finally, we created the remaining 90 databases based on WikiSQL. To ensure the domain diversity, we select about 500 tables in about 90 different domains to create these 90 databases. To create each database, we chose several related tables from WikiSQL dev or test splits, and then created a relational database schema with foreign keys based on the tables we selected. We had to create some intersection tables in order to link several tables together. For most other cases, we did not need to populate these databases since tables in WikiSQL are from Wikipedia, which already had real world data stored.

We manually corrected some database schemas if they had some column names that did not make sense or missed some foreign keys. For table and column names, it is common to use abbreviations in databases. For example, `student\_id' might be represented by `stu\_id'. For our task definition, we manually changed each column name back to regular words so that the system only handled semantic parsing issues. 

\subsection{Question and SQL Annotation}
\label{sec:question_and_sql_annnotation}

For each database, we ask eight computer science students proficient in SQL to create 20-50 natural questions and their SQL labels. 
To make our questions diverse, natural, and reflective of how humans actually use databases, 
we did not use any template or script to generate question and SQL queries.
Our annotation procedure ensures the following three aspects.
 
\paragraph{A)~ SQL pattern coverage.}
We ensure that our corpus contains enough examples for all common SQL patterns. For each database, we ask annotators to write SQL queries that cover all the following SQL components: \texttt{SELECT} with multiple columns and aggregations, \texttt{WHERE}, \texttt{GROUP BY}, \texttt{HAVING}, \texttt{ORDER BY}, \texttt{LIMIT}, \texttt{JOIN}, \texttt{INTERSECT}, \texttt{EXCEPT}, \texttt{UNION}, \texttt{NOT IN}, \texttt{OR}, \texttt{AND}, \texttt{EXISTS}, \texttt{LIKE} as well as nested queries. The annotators made sure that each table in the database appears in at least one query.



\paragraph{B)~ SQL consistency.}
Some questions have multiple acceptable SQL queries with the same result.
However, giving totally different SQL labels to similar questions can hinder the training of semantic parsing models.
To avoid this issue, we designed the annotation protocol so that
all annotators choose the same SQL query pattern if multiple equivalent queries are possible. 

\paragraph{C)~ Question clarity.}
We did not create questions that are (1) vague or too ambiguous, or (2) require knowledge outside the database to answer.

First, ambiguous questions refer to the questions that do not have enough clues to infer which columns to return and which conditions to consider. 
For example, we would not ask ``What is the most popular class at University X?'' because the definition of ``popular'' is not clear: it could mean the rating of the class or the number of students taking the course.
Instead, we choose to ask ``What is the name of the class which the largest number of students are taking at University X?''. Here, ``popular'' refers to the size of student enrollment. Thus, the ``student\_enrollment'' column can be used in condition to answer this question.
We recognize that ambiguous questions appear in real-world natural language database interfaces. 

We agree that future work needs to address this issue by having multi-turn interactions between the system and users for clarification. However, our main aim here is to develop a corpus to tackle the problem of handling complex queries and generalizing across databases without multi-turn interactions required, which no existing semantic parsing datasets could do. Moreover, the low performances of current state-of-the-art models already show that our task is challenging enough, without ambiguous questions.
In addition, questions are required to contain the specific information to return.
Otherwise, we don't know if class id is also acceptable in the previous case. Most of the questions in the existing semantic parsing datasets are ambiguous. This is not a serious problem if we use one single dataset because we have enough data domain specific examples to know which columns are the default.
However, it would be a serious problem in cross domain tasks since the default return values differ cross domain and people.

Second, humans sometimes ask questions that require common sense knowledge outside the given database.
For instance, when people ask ``Display the employee id for the employees who report to John", the correct SQL is
\begin{quote}
\texttt{SELECT employee\_id \\ FROM employees \\ WHERE manager\_id = (\\ SELECT employee\_id \\ FROM employees \\ \quad WHERE first\_name = `John')}
\end{quote}
which requires the common knowledge that ``X reports to Y" corresponds to an ``employee-manager" relation.
we do not include such questions and leave them as a future research direction.


\paragraph{Annotation tools} We open each database on a web-based interface powered by the sqlite\_web\footnote{\url{https://github.com/coleifer/sqlite-web}} tool. It allows the annotators to see the schema and content of each table, execute SQL queries, and check the returned results. 
This tool was extremely helpful for the annotators to write executable SQL queries that reflect the true meaning of the given questions and return correct answers.

\begin{table*}[ht!]
\centering
\scalebox{0.78}{
\begin{tabular}{c|ccccccccc}
\hline
Dataset & \# Q  & \# SQL & \# DB & \# Domain & \!\!\# Table \scalebox{0.7}[0.8]{/\,DB}\!\! & \texttt{ORDER BY} & \texttt{GROUP BY} & \texttt{NESTED} & \texttt{HAVING} \\ \hline
ATIS & 5,280  & 947  & 1 & 1 & 32 & 0 & 5 & 315 & 0 \\
GeoQuery & 877   & 247  & 1 & 1 & 6 & 20 & 46 & 167 & 9\\
Scholar  & 817   & 193  & 1 & 1 & 7  & 75 & 100 & 7 & 20\\
Academic & 196  & 185  & 1 & 1 & 15  & 23 & 40 & 7 & 18\\
IMDB & 131  & 89  & 1 & 1 & 16 & 10 & 6 & 1 & 0\\
Yelp & 128  & 110  & 1 & 1 & 7 & 18 & 21 & 0 & 4\\
Advising & 3,898  & 208 & 1 & 1 & 10 & 15 & 9 & 22 & 0\\
Restaurants & 378 & 378 & 1 & 1 & 3 & 0 & 0 & 4 & 0\\
WikiSQL  & 80,654 & 77,840 & 26,521 & - & 1 & 0 & 0 & 0 & 0\\ \hline
\textbf{Spider} & 10,181 & 5,693 & 200 & ~138 & 5.1 & 1335 & 1491 & 844 & 388\\ \hline
\end{tabular}}
\vspace{-1mm}
\caption{Comparisons of text-to-SQL datasets. \textbf{Spider} is the \textit{only one} text-to-SQL dataset that contains both databases with multiple tables in different domains and complex SQL queries. It was designed to test the ability of a system to generalize to not only new SQL queries and database schemas but also new domains.} 
\label{tb:data}
\vspace{-2mm}
\end{table*}

\subsection{SQL Review}
\label{sec:sql_review}

Once the database is labeled with question-query pairs, we ask a different annotator to check if the questions are clear and contain enough information to answer the query. 
For a question with multiple possible SQL translations, the reviewers double check whether the SQL label is correctly chosen under our protocol. 
Finally, the reviewers check if all the SQL labels in the current database cover all the common SQL clauses.

\subsection{Question Review and Paraphrase}
\label{sec:question_review}

After SQL labels are reviewed, native English speakers review and correct each question. 
They first check if the question is grammatically correct and natural. 
Next, they make sure that the question reflects the meaning of its corresponding SQL label. Finally, to 
improve the diversity in questions, we ask annotators to add a paraphrased version to some questions.

\subsection{Final Review}
\label{sec:final_question}

Finally, we ask the most experienced annotator to conduct the final question and SQL review. This annotator makes the final decision if multiple reviewers are not sure about some annotation issues. Also, we run a script to execute and parse all SQL labels to make sure they are correct.

\section{Dataset Statistics and Comparison}
\label{sec:data_analysis}
We summarize the statistics of Spider and other text-to-SQL datasets in Table \ref{tb:data}.
Compared with other datasets, Spider contains databases with multiple tables and contains SQL queries including many complex SQL components. 
For example, Spider contains about twice more nested queries and 10 times more \texttt{ORDER BY (LIMIT)} and \texttt{GROUP BY (HAVING)} components
than the total of previous text-to-SQL datasets.
Spider has 200 distinct databases covering 138 different domains such as college, club, TV show, government, etc.
Most domains have one database, thus containing 20-50 questions, and a few domains such as flight information have multiple databases with more than 100 questions in total.
On average, each database in Spider has 27.6 columns and 8.8 foreign keys.
The average question length and SQL length are about 13 and 21 respectively.
Our task uses different databases for training and testing, evaluating the cross-domain performance.
Therefore, Spider is the \textit{only one} text-to-SQL dataset that contains both databases with multiple tables in different domains and complex SQL queries
It tests the ability of a system to generalize to not only new SQL queries and database schemas but also new domains.




\section{Task Definition}
\label{sec:task}
On top of the proposed dataset, we define a text-to-SQL task that is more realistic than prior work. Unlike most of the previous semantic parsing or text-to-SQL tasks, models will be tested on \textit{both different complex SQL queries and different complex databases in different domains} in our task. It aims to ensure that models can only make the correct prediction when they truly understand the semantic meaning of the questions, rather than just memorization. Also, because our databases contain different domains, our corpus tests model's ability to generalize to new databases. In this way, model performance on this task can reflect the real semantic parsing ability.

In order to make the task feasible and to focus on the more fundamental part of semantic parsing, we make the following assumptions:

\begin{itemize}
\setlength{\leftskip}{-3mm}
    \item In our current task, we do not evaluate model performance on generating values. Predicting correct SQL structures and columns is more realistic and critical at this stage based on the low performances of various current state-of-the-art models on our task. 
    In a real world situation, people need to double check what condition values are and finalize them after multiple times. It is unrealistic to predict condition values without interacting with users. In reality, most people know what values to ask but do not know the SQL logic. A more reasonable way is to ask users to use an interface searching the values, then ask more specific questions. Also, other previous work with value prediction uses one single database in both train and test which makes it vulnerable to overfitting. However, SQL queries must include values in order to execute them. For value prediction in our task, a list of gold values for each question is given. Models need to fill them into the right slots in their predicted SQL.
    
    \item As mentioned in the previous sections, we exclude some queries that require outside knowledge such as common sense inference and math calculation. For example, imagine a table with birth and death year columns. To answer the questions like ``How long is X's life length?'', we use \texttt{SELECT death\_year - birth\_year}. Even though this example is easy for humans, it requires some common knowledge of the life length definition and the use of a math operation, which is not the focus of our dataset.
    
    \item We assume all table and column names in the database are clear and self-contained.
    For example, some databases use database specific short-cut names for table and column names such as ``stu\_id'', which we manually converted to ``student id'' in our corpus.
    
\end{itemize}

\section{Evaluation Metrics}
\label{sec:eval}

Our evaluation metrics include Component Matching, Exact Matching, and Execution Accuracy.
In addition, we measure the system's accuracy as a function of the difficulty of a query.
Since our task definition does not predict value string, our evaluation metrics do not take value strings into account.

We will release the official evaluation script along with our corpus so that the research community can share the same evaluation platform.

\paragraph{Component Matching} 
To conduct a detailed analysis of model performance, we measure the average exact match between the prediction and ground truth on different SQL components. For each of the following components: \vspace{-2mm}
\begin{itemize}
    \setlength{\itemsep}{-1mm}
    \item \texttt{SELECT} \quad $\bullet$ \texttt{WHERE} \quad $\bullet$ \texttt{GROUP BY}
    \item  \texttt{ORDER BY} \quad $\bullet$ \texttt{KEYWORDS} (including all SQL keywords without column names and operators)
\end{itemize}\vspace{-2mm}
we decompose each component in the prediction and the ground truth as bags of several sub-components, and check whether or not these two sets of components match exactly.
To evaluate each \texttt{SELECT} component, for example, consider \texttt{SELECT avg(col1), max(col2), min(col1)},  we first parse and decompose into a set \texttt{(avg, min, col1), (max, col2)}, and see if the gold and predicted sets are the same.
Previous work directly compared decoded SQL with gold SQL. 
However, some SQL components do not have order constraints.
In our evaluation, we treat each component as a set so that for example, \texttt{SELECT avg(col1),  min(col1), max(col2)} and \texttt{SELECT avg(col1), max(col2), min(col1)} would be treated as the same query.
To report a model's overall performance on each component, we compute F1 score on exact set matching.

\paragraph{Exact Matching} 
We measure whether the predicted query as a whole is equivalent to the gold query.
We first evaluate the SQL clauses as described in the last section. The predicted query is correct only if all of the components are correct. Because we conduct a set comparison in each clause, this exact matching metric can handle the ``ordering issue'' \cite{Xu2017}.

\paragraph{Execution Accuracy\footnote{Please check our website for the latest updates on the task at \url{https://yale-lily.github.io/spider}}}
Since Exact Matching is possible to provide false negative evaluation when the semantic parser is able to generate novel syntax structures, we also consider Execution Accuracy.
For Execution Accuracy, the value is a must in order to execute SQL queries. Instead of generating these values, a list of gold values for each question is given. Models need to select them and fill them into the right slots in their predicted SQL.
We exclude value prediction in Component and Exact Matching evaluations and do not provide Execution Accuracy in the current version. 
However, it is also important to note that Execution Accuracy can create false positive evaluation when a predicted SQL returns the same result (for example, `NULL') as the gold SQL while they are semantically different.
So we can use both to complement each other.

Finally, our evaluation also considers multiple acceptable keys if \texttt{JOIN} and \texttt{GROUP} are in the query. For example, suppose ``stu\_id'' in one table refers to ``stu\_id'' in another table, \texttt{GROUP BY} either is acceptable.

\begin{figure}[!t]
    \centering
    \includegraphics[width=0.5\textwidth]{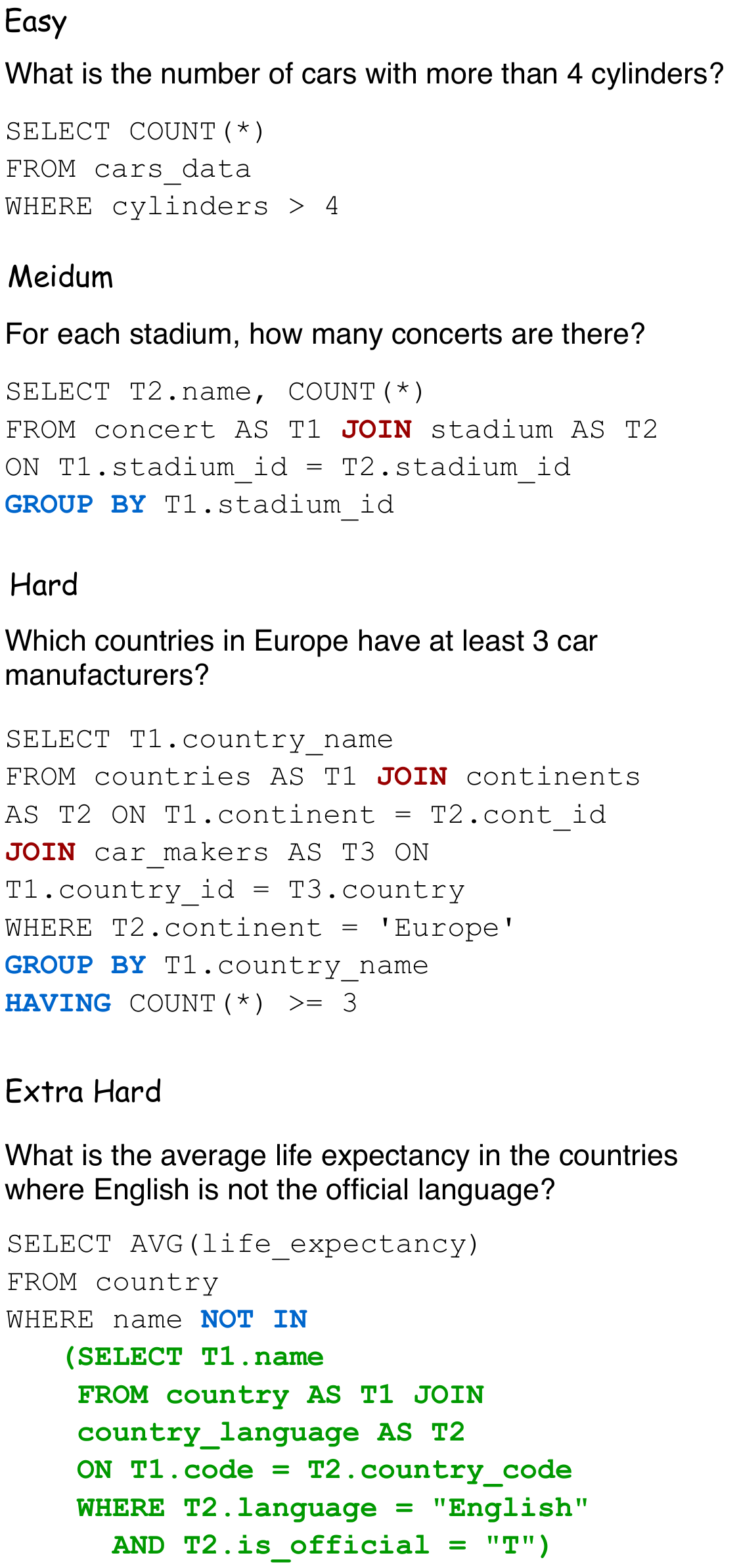}
    \caption{SQL query examples in 4 hardness levels.}
\label{fig:example}
\end{figure}

\paragraph{SQL Hardness Criteria} 
To better understand the model performance on different queries, we divide SQL queries into 4 levels: easy, medium, hard, extra hard.
We define the difficulty based on the number of SQL components, selections, and conditions, so that queries that contain more SQL keywords (\texttt{GROUP BY}, \texttt{ORDER BY}, \texttt{INTERSECT}, nested subqueries, column selections and aggregators, etc) are considered to be harder.
For example, a query is considered as hard if it includes more than two \texttt{SELECT} columns, more than two \texttt{WHERE} conditions, and \texttt{GROUP BY} two columns, or contains \texttt{EXCEPT} or nested queries. A SQL with more additions on top of that is considered as extra hard.
Figure \ref{fig:example} shows examples of SQL queries in 4 hardness levels.

\begin{table*}[ht!]
\centering
\scalebox{0.92}{
\begin{tabular}{c|ccccc|c}
\hline
                                & \multicolumn{5}{c|}{Test} & Dev \\
                          & Easy    & Medium  & Hard   & Extra Hard & All & All   \\ \hline
                                \multicolumn{6}{c}{Example Split} \\ \hline
Seq2Seq                         & 22.0 & 7.8 & 5.5 & 1.3 & 9.4 & 10.3  \\
Seq2Seq+Attention \cite{dong16} & 32.3 & 15.6 & 10.3 & 2.3 & 15.9 & 16.0   \\
Seq2Seq+Copying                 & 29.3 & 13.1 & 8.8 & 3.0 & 14.1 & 15.3  \\
SQLNet \cite{Xu2017}            & 34.1 & 19.6 & 11.7 & 3.3 & 18.3 & 18.4   \\
TypeSQL \cite{Yu18}             & 47.5 & 38.4 & 24.1 & 14.4 & 33.0 & 34.4   \\ \hline
                                \multicolumn{6}{c}{Database Split} \\ \hline
Seq2Seq                         & 11.9 & 1.9 & 1.3 & 0.5 & 3.7 & 1.9  \\
Seq2Seq+Attention \cite{dong16} & 14.9 & 2.5 & 2.0 & 1.1 & 4.8 & 1.8   \\
Seq2Seq+Copying                 & 15.4 & 3.4 & 2.0 & 1.1 & 5.3 & 4.1  \\
SQLNet \cite{Xu2017}            & 26.2 & 12.6 & 6.6 & 1.3 & 12.4 & 10.9   \\
TypeSQL \cite{Yu18}             & 19.6 & 7.6 & 3.8 & 0.8 & 8.2 & 8.0   \\ \hline   
\end{tabular}}
\vspace{-2mm}
\caption{Accuracy of Exact Matching on SQL queries with different hardness levels.}
\label{tab:results}
\vspace{-1mm}
\end{table*}

\begin{table*}[ht!]
\centering
\scalebox{0.92}{
\begin{tabular}{ccccccc}
\hline
 Method           & \texttt{SELECT} & \texttt{WHERE} & \texttt{GROUP BY} & \texttt{ORDER BY} & \texttt{KEYWORDS} \\\hline
                  \multicolumn{6}{c}{Example Split} \\ \hline
Seq2Seq           & 23.3 & 4.9 & 15.3 & 9.2 & 17.9  \\
Seq2Seq+Attention & 31.1 & 9.1 & 28.2 & 20.8 & 21.4  \\
Seq2Seq+Copying   & 28.2 & 8.3 & 25.5 & 21.3 & 19.0  \\
SQLNet            & 59.8 & 32.9 & 35.9 & 65.5 & 76.1  \\
TypeSQL           & 77.3 & 52.4 & 47.0 & 67.5 & 78.4  \\ \hline
                  \multicolumn{6}{c}{Database Split} \\ \hline
Seq2Seq           & 13.0 & 1.5 & 3.3 & 5.3 & 8.7  \\
Seq2Seq+Attention & 13.6 & 3.1 & 3.6 & 9.9 & 9.9  \\
Seq2Seq+Copying   & 12.0 & 3.1 & 5.3 & 5.8 & 7.3  \\
SQLNet            & 44.5 & 19.8 & 29.5 & 48.8 & 64.0 \\
TypeSQL           & 36.4 & 16.0 & 17.2 & 47.7 & 66.2 \\ \hline
\end{tabular}}
\vspace{-2mm}
\caption{F1 scores of Component Matching on all SQL queries on Test set.}
\label{tab:results_component}
\vspace{-3mm}
\end{table*}

\section{Methods}
\label{sec:methods}
In order to analyze the difficulty and demonstrate the purpose of our corpus, we experiment with several state-of-the-art semantic parsing models.
As our dataset is fundamentally different from the prior datasets such as Geoquery and WikiSQL, we adapted these models to our task as follows.
We created a `big' column list by concatenating columns in all tables of the database together as an input to all models.  Also, for each model, we limit the column selection space for each question example to all column of the database which the question is asking instead of all column names in the whole corpus.

\paragraph{Seq2Seq}
Inspired by neural machine translation \cite{sutskever2014sequence}, we first apply a basic sequence-to-sequence model, \textbf{Seq2Seq}.
Then, we also explore \textbf{Seq2Seq+Attention} from \cite{dong16} by adding an attention mechanism \cite{bahdanau2015neural}.
In addition, we include \textbf{Seq2Seq+Copying} by adding an attention-based copying operation similar to \cite{jia2016}.

The original model does not take the schema into account because it has the same schema in both train and test. We modify the model so that it considers the table schema information by passing a vocabulary mask that limits the model to decode the words from SQL keywords, table and column names in the current database.

\paragraph{SQLNet}
introduced by \cite{Xu2017} uses column attention and employs a sketch-based method and generates SQL as a slot-filling task.
This fundamentally avoids the sequence-to-sequence structure when ordering does not matter in SQL query conditions.
Because it is originally designed for WikiSQL, we extend its \texttt{SELECT} and \texttt{WHERE} modules to \texttt{ORDER BY} and \texttt{GROUP BY} components.
 
\paragraph{TypeSQL} proposed by \cite{Yu18} improves upon SQLNet by proposing a different training procedure and utilizing types extracted from either knowledge graph or table content to help model better understand entities and numbers in the question.
In our experiment, we use the question type info extracted from database content. Also, we extend their modules to \texttt{ORDER BY} and \texttt{GROUP BY} components as well. It is the only model that uses database content.

\section{Experimental Results and Discussion}
We summarize the performance of all models on our test set including accuracy of exact matching in Table \ref{tab:results} and F1 scores of component matching in Table \ref{tab:results_component}.
\paragraph{Data Splits} 
For the final training dataset, we also select and include 752 queries and 1659 questions that follow our annotation protocol from six existing datasets: Restaurants, GeoQuery, Scholar, Academic, IMDB, and Yelp.
We report results on two different settings for all models: (1) Example split where examples are randomly split into 8659 train, 1034 dev, 2147 test. Questions for the same database can appear in both train and test. (2) Database split where 206 databases are split into 146 train, 20 dev, and 40 test. All questions for the same database are in the same split.
\paragraph{Overall Performance} 
The performances of the Seq2Seq-based basic models including Seq2Seq, Seq2Seq+Attention, and Seq2Seq+Copying are very low. However, they are able to generate nested and complex queries because of their general decoding process.
Thus, they can get a few hard and extra hard examples correct.
But in the vast majority of cases, they predict invalid SQL queries with grammatical errors.
The attention and copying mechanisms do not help much either.

In contrast, SQLNet and TypeSQL that utilize SQL structure information to guide the SQL generation process significantly outperform other Seq2Seq models. While they can produce valid queries, however, they are unable to generate nested queries or queries with keywords such as \texttt{EXCEPT} and \texttt{INTERSECT} because they limit possible SQL outputs in some fixed pre-defined SQL structures.

As Component Matching results in Table \ref{tab:results_component} shows, all models struggle with \texttt{WHERE} clause prediction the most. \texttt{WHERE} clause is more likely to have multiple columns and operations, which makes its prediction the most challenging. The most number of prediction errors for each component is from column prediction.

In general, the overall performances of all models are low, indicating that our task is challenging and there is still a large room for improvement.
\paragraph{Example Split vs Database Split}
As discussed in Section \ref{sec:task}, another challenge of the dataset is to generalize to new databases.
To study this, in Table \ref{tab:results} and Table \ref{tab:results_component} we compare model performances under the two settings.
For all models, the performance under database split is much lower than that under example split.
Especially, TypeSQL utilizes column names as question types, and it outperforms other models with a large margin under the example split. However, its performance drops the most on the database split data set. This indicates that the model does well on complex SQL prediction but fails to generalize to new databases.
In addition, we observe that all models perform much poorer on column selection under database split than example split. 

Overall, the result shows that our dataset presents a challenge for the model to generalize to new databases.

\begin{figure}[!t]
    \vspace{-1.5mm}\hspace{-1mm}
    \centering
    \includegraphics[width=0.48\textwidth]{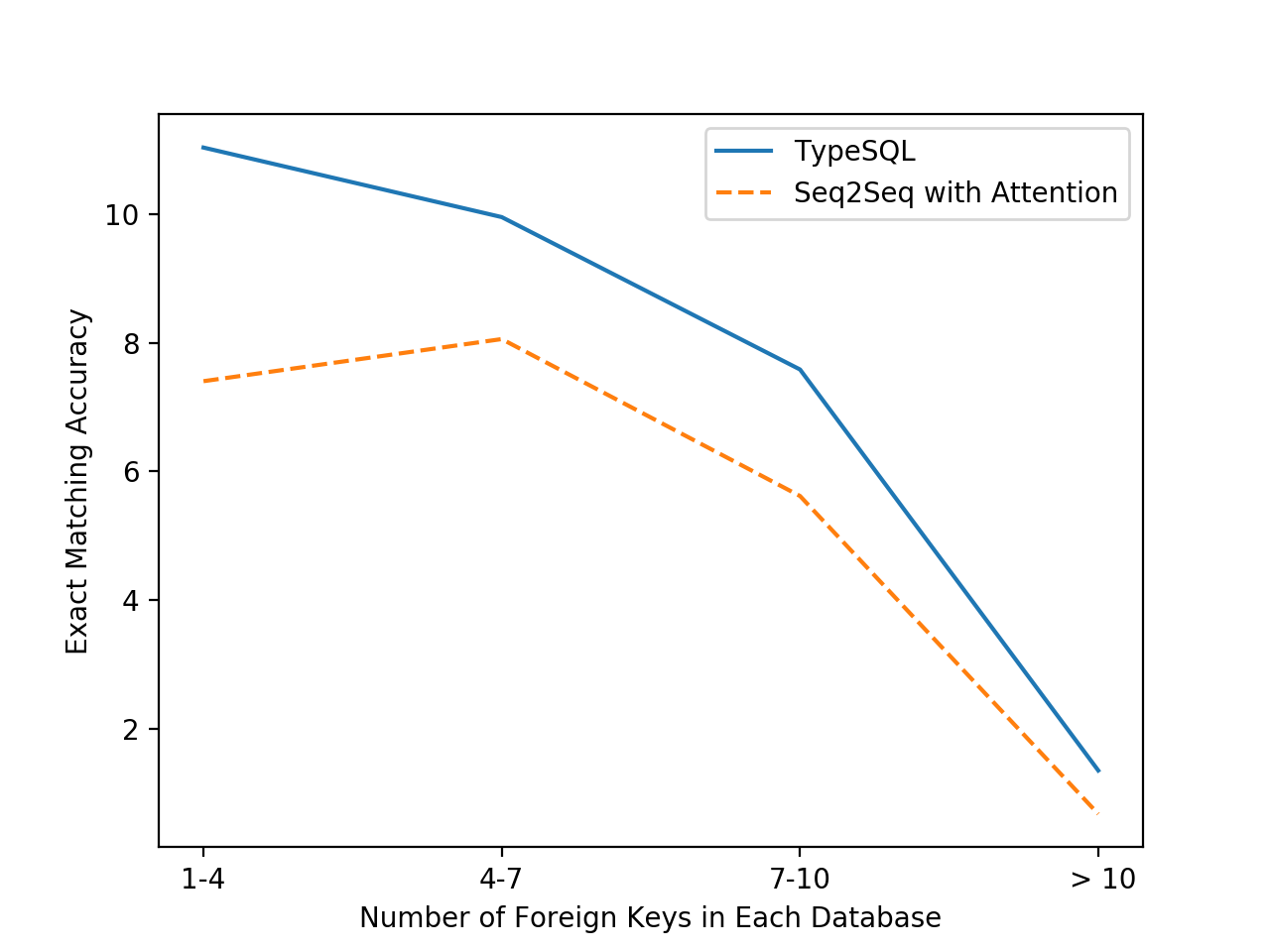}\vspace{-3mm}
    \caption{Exact matching accuracy as a function of the number of foreign keys.}
\label{fig:complexity}
\vspace{-4mm}
\end{figure}
\paragraph{Complexity of Database Schema}
In order to show how the complexity of the database schema affects model performance, Figure \ref{fig:complexity} plots the exact matching accuracy as a function of the number of foreign keys in a database. 
The performance decreases as the database has more foreign keys.
The first reason is that the model has to choose column and table names from many candidates in a complex database schema.
Second, a complex database schema presents a great challenge for the model to capture the relationship between different tables with foreign keys.
SQL answers to questions on the database with more number of foreign keys are more likely to join more tables.
It indicates that this task requires more effective methods to encode the relation of tables with foreign keys.

\section{Conclusion}
\label{sec:conclusion}

In this paper we introduce Spider, a large, complex and cross-domain semantic parsing and text-to-SQL dataset,
which directly benefits both NLP and DB communities.
Based on Spider, we define a 
new challenging and realistic semantic parsing task.
Experimental results on several state-of-the-art models on this task suggest plenty space for improvement.

\section*{Acknowledgement}
We thank Graham Neubig, Tianze Shi, Catherine Finegan-Dollak, and three anonymous reviewers for their discussion and feedback.
We also thank Barry Williams for providing part of our database schemas from the DatabaseAnswers.

\bibliography{emnlp2018}

\begin{thebibliography}{46}
\expandafter\ifx\csname natexlab\endcsname\relax\def\natexlab#1{#1}\fi

\bibitem[{Allamanis et~al.(2015)Allamanis, Tarlow, Gordon, and
  Wei}]{Allamanis15}
Miltiadis Allamanis, Daniel Tarlow, Andrew~D. Gordon, and Yi~Wei. 2015.
\newblock Bimodal modelling of source code and natural language.
\newblock In \emph{{ICML}}, volume~37 of \emph{{JMLR} Workshop and Conference
  Proceedings}, pages 2123--2132. JMLR.org.

\bibitem[{Artzi and Zettlemoyer(2013)}]{artzi13}
Yoav Artzi and Luke Zettlemoyer. 2013.
\newblock Weakly supervised learning of semantic parsers for mapping
  instructions to actions.
\newblock \emph{Transactions of the Association forComputational Linguistics}.

\bibitem[{Bahdanau et~al.(2015)Bahdanau, Cho, and Bengio}]{bahdanau2015neural}
Dzmitry Bahdanau, Kyunghyun Cho, and Yoshua Bengio. 2015.
\newblock Neural machine translation by jointly learning to align and
  translate.
\newblock In \emph{ICLR}.

\bibitem[{Banarescu et~al.(2013)Banarescu, Bonial, Cai, Georgescu, Griffitt,
  Hermjakob, Knight, Koehn, Palmer, and Schneider}]{banarescu13}
Laura Banarescu, Claire Bonial, Shu Cai, Madalina Georgescu, Kira Griffitt, Ulf
  Hermjakob, Kevin Knight, Philipp Koehn, Martha Palmer, and Nathan Schneider.
  2013.
\newblock Abstract meaning representation for sembanking.
\newblock In \emph{Proceedings of the 7th Linguistic Annotation Workshop and
  Interoperability with Discourse}.

\bibitem[{Berant and Liang(2014)}]{Berant14}
Jonathan Berant and Percy Liang. 2014.
\newblock Semantic parsing via paraphrasing.
\newblock In \emph{Proceedings of the 52nd Annual Meeting of the Association
  for Computational Linguistics (Volume 1: Long Papers)}, pages 1415--1425,
  Baltimore, Maryland. Association for Computational Linguistics.

\bibitem[{Dahl et~al.(1994)Dahl, Bates, Brown, Fisher, Hunicke-Smith, Pallett,
  Pao, Rudnicky, and Shriberg}]{Dahl94}
Deborah~A. Dahl, Madeleine Bates, Michael Brown, William Fisher, Kate
  Hunicke-Smith, David Pallett, Christine Pao, Alexander Rudnicky, and
  Elizabeth Shriberg. 1994.
\newblock Expanding the scope of the atis task: The atis-3 corpus.
\newblock In \emph{Proceedings of the Workshop on Human Language Technology},
  HLT '94, Stroudsburg, PA, USA. Association for Computational Linguistics.

\bibitem[{Das et~al.(2010)Das, Schneider, Chen, and Smith}]{Das10}
Dipanjan Das, Nathan Schneider, Desai Chen, and Noah~A. Smith. 2010.
\newblock Probabilistic frame-semantic parsing.
\newblock In \emph{NAACL}.

\bibitem[{Deng et~al.(2009)Deng, Dong, Socher, Li, Li, and
  Fei-Fei}]{imagenet_cvpr09}
J.~Deng, W.~Dong, R.~Socher, L.-J. Li, K.~Li, and L.~Fei-Fei. 2009.
\newblock {ImageNet: A Large-Scale Hierarchical Image Database}.
\newblock In \emph{CVPR09}.

\bibitem[{Dong and Lapata(2016)}]{dong16}
Li~Dong and Mirella Lapata. 2016.
\newblock Language to logical form with neural attention.
\newblock In \emph{Proceedings of the 54th Annual Meeting of the Association
  for Computational Linguistics, {ACL} 2016, August 7-12, 2016, Berlin,
  Germany, Volume 1: Long Papers}.

\bibitem[{Dong and Lapata(2018)}]{P18-1068}
Li~Dong and Mirella Lapata. 2018.
\newblock Coarse-to-fine decoding for neural semantic parsing.
\newblock In \emph{Proceedings of the 56th Annual Meeting of the Association
  for Computational Linguistics (Volume 1: Long Papers)}, pages 731--742.
  Association for Computational Linguistics.

\bibitem[{Finegan-Dollak et~al.(2018)Finegan-Dollak, Kummerfeld, Zhang,
  Ramanathan, Sadasivam, Zhang, and Radev}]{cathy18}
Catherine Finegan-Dollak, Jonathan~K. Kummerfeld, Li~Zhang, Karthik
  Ramanathan~Dhanalakshmi Ramanathan, Sesh Sadasivam, Rui Zhang, and Dragomir
  Radev. 2018.
\newblock Improving text-to-sql evaluation methodology.
\newblock In \emph{ACL 2018}. Association for Computational Linguistics.

\bibitem[{Giordani and Moschitti(2012)}]{giordani2012translating}
Alessandra Giordani and Alessandro Moschitti. 2012.
\newblock Translating questions to sql queries with generative parsers
  discriminatively reranked.
\newblock In \emph{COLING (Posters)}, pages 401--410.

\bibitem[{Huang et~al.(2018)Huang, Wang, Singh, tau Yih, and
  He}]{pshuang2018PT-MAML}
Po{-}Sen Huang, Chenglong Wang, Rishabh Singh, Wen tau Yih, and Xiaodong He.
  2018.
\newblock Natural language to structured query generation via meta-learning.
\newblock In \emph{NAACL}.

\bibitem[{Iyer et~al.(2017)Iyer, Konstas, Cheung, Krishnamurthy, and
  Zettlemoyer}]{iyer17}
Srinivasan Iyer, Ioannis Konstas, Alvin Cheung, Jayant Krishnamurthy, and Luke
  Zettlemoyer. 2017.
\newblock Learning a neural semantic parser from user feedback.
\newblock \emph{CoRR}, abs/1704.08760.

\bibitem[{Jia and Liang(2016)}]{jia2016}
Robin Jia and Percy Liang. 2016.
\newblock Data recombination for neural semantic parsing.
\newblock In \emph{Proceedings of the 54th Annual Meeting of the Association
  for Computational Linguistics (Volume 1: Long Papers)}.

\bibitem[{Li and Jagadish(2014)}]{li2014constructing}
Fei Li and HV~Jagadish. 2014.
\newblock Constructing an interactive natural language interface for relational
  databases.
\newblock \emph{VLDB}.

\bibitem[{Li et~al.(2006)Li, Yang, and Jagadish}]{li2006constructing}
Yunyao Li, Huahai Yang, and HV~Jagadish. 2006.
\newblock Constructing a generic natural language interface for an xml
  database.
\newblock In \emph{EDBT}, volume 3896, pages 737--754. Springer.

\bibitem[{Liang et~al.(2011)Liang, Jordan, and Klein}]{Liang11}
P.~Liang, M.~I. Jordan, and D.~Klein. 2011.
\newblock Learning dependency-based compositional semantics.
\newblock In \emph{Association for Computational Linguistics (ACL)}, pages
  590--599.

\bibitem[{Lin et~al.(2018)Lin, Wang, Zettlemoyer, and Ernst}]{nl2bash}
Xi~Victoria Lin, Chenglong Wang, Luke Zettlemoyer, and Michael~D. Ernst. 2018.
\newblock Nl2bash: A corpus and semantic parser for natural language interface
  to the linux operating system.
\newblock In \emph{LREC}.

\bibitem[{Ling et~al.(2016)Ling, Blunsom, Grefenstette, Hermann, Kocisk{\'{y}},
  Wang, and Senior}]{ling16}
Wang Ling, Phil Blunsom, Edward Grefenstette, Karl~Moritz Hermann, Tom{\'{a}}s
  Kocisk{\'{y}}, Fumin Wang, and Andrew Senior. 2016.
\newblock Latent predictor networks for code generation.
\newblock In \emph{{ACL} {(1)}}. The Association for Computer Linguistics.

\bibitem[{McCann et~al.(2018)McCann, Keskar, Xiong, and
  Socher}]{mccann2018natural}
Bryan McCann, Nitish~Shirish Keskar, Caiming Xiong, and Richard Socher. 2018.
\newblock The natural language decathlon: Multitask learning as question
  answering.
\newblock \emph{arXiv preprint arXiv:1806.08730}.

\bibitem[{Oda et~al.(2015)Oda, Fudaba, Neubig, Hata, Sakti, Toda, and
  Nakamura}]{Oda15}
Yusuke Oda, Hiroyuki Fudaba, Graham Neubig, Hideaki Hata, Sakriani Sakti,
  Tomoki Toda, and Satoshi Nakamura. 2015.
\newblock Learning to generate pseudo-code from source code using statistical
  machine translation (t).
\newblock In \emph{Proceedings of the 2015 30th IEEE/ACM International
  Conference on Automated Software Engineering (ASE)}, ASE '15.

\bibitem[{Popescu et~al.(2004)Popescu, Armanasu, Etzioni, Ko, and
  Yates}]{popescu2004modern}
Ana-Maria Popescu, Alex Armanasu, Oren Etzioni, David Ko, and Alexander Yates.
  2004.
\newblock Modern natural language interfaces to databases: Composing
  statistical parsing with semantic tractability.
\newblock In \emph{Proceedings of the 20th international conference on
  Computational Linguistics}, page 141. Association for Computational
  Linguistics.

\bibitem[{Popescu et~al.(2003{\natexlab{a}})Popescu, Etzioni, and
  Kautz}]{Popescu03}
Ana-Maria Popescu, Oren Etzioni, and Henry Kautz. 2003{\natexlab{a}}.
\newblock Towards a theory of natural language interfaces to databases.
\newblock In \emph{Proceedings of the 8th International Conference on
  Intelligent User Interfaces}.

\bibitem[{Popescu et~al.(2003{\natexlab{b}})Popescu, Etzioni, and
  Kautz}]{popescu2003towards}
Ana-Maria Popescu, Oren Etzioni, and Henry Kautz. 2003{\natexlab{b}}.
\newblock Towards a theory of natural language interfaces to databases.
\newblock In \emph{Proceedings of the 8th international conference on
  Intelligent user interfaces}, pages 149--157. ACM.

\bibitem[{Price(1990)}]{Price90}
P.~J. Price. 1990.
\newblock Evaluation of spoken language systems: the atis domain.
\newblock In \emph{Speech and Natural Language: Proceedings of a Workshop Held
  at Hidden Valley, Pennsylvania, June 24-27,1990}, pages 91--95.

\bibitem[{Quirk et~al.(2015)Quirk, Mooney, and Galley}]{quirk2015language}
Chris Quirk, Raymond Mooney, and Michel Galley. 2015.
\newblock Language to code: Learning semantic parsers for if-this-then-that
  recipes.
\newblock In \emph{Proceedings of the 53rd Annual Meeting of the Association
  for Computational Linguistics and the 7th International Joint Conference on
  Natural Language Processing (Volume 1: Long Papers)}, volume~1, pages
  878--888.

\bibitem[{Rabinovich et~al.(2017)Rabinovich, Stern, and Klein}]{RabinovichSK17}
Maxim Rabinovich, Mitchell Stern, and Dan Klein. 2017.
\newblock Abstract syntax networks for code generation and semantic parsing.
\newblock In \emph{{ACL} {(1)}}, pages 1139--1149. Association for
  Computational Linguistics.

\bibitem[{Rajpurkar et~al.(2016)Rajpurkar, Zhang, Lopyrev, and
  Liang}]{Pranav16}
Pranav Rajpurkar, Jian Zhang, Konstantin Lopyrev, and Percy Liang. 2016.
\newblock Squad: 100, 000+ questions for machine comprehension of text.
\newblock \emph{CoRR}, abs/1606.05250.

\bibitem[{Reddy et~al.(2014)Reddy, Lapata, and Steedman}]{Reddy14}
Siva Reddy, Mirella Lapata, and Mark Steedman. 2014.
\newblock Large-scale semantic parsing without question-answer pairs.
\newblock \emph{Transactions of the Association for Computational Linguistics},
  2:377--392.

\bibitem[{Sutskever et~al.(2014)Sutskever, Vinyals, and
  Le}]{sutskever2014sequence}
Ilya Sutskever, Oriol Vinyals, and Quoc~V Le. 2014.
\newblock Sequence to sequence learning with neural networks.
\newblock In \emph{Advances in neural information processing systems}, pages
  3104--3112.

\bibitem[{Tang and Mooney(2001{\natexlab{a}})}]{tang01}
Lappoon~R. Tang and Raymond~J. Mooney. 2001{\natexlab{a}}.
\newblock Using multiple clause constructors in inductive logic programming for
  semantic parsing.
\newblock In \emph{Proceedings of the 12th European Conference on Machine
  Learning}, pages 466--477, Freiburg, Germany.

\bibitem[{Tang and Mooney(2001{\natexlab{b}})}]{tang2001using}
Lappoon~R Tang and Raymond~J Mooney. 2001{\natexlab{b}}.
\newblock Using multiple clause constructors in inductive logic programming for
  semantic parsing.
\newblock In \emph{ECML}, volume~1, pages 466--477. Springer.

\bibitem[{Wang et~al.(2017)Wang, Cheung, and Bodik}]{wang2017synthesizing}
Chenglong Wang, Alvin Cheung, and Rastislav Bodik. 2017.
\newblock Synthesizing highly expressive sql queries from input-output
  examples.
\newblock In \emph{Proceedings of the 38th ACM SIGPLAN Conference on
  Programming Language Design and Implementation}, pages 452--466. ACM.

\bibitem[{Wang et~al.(2018)Wang, Huang, Polozov, Brockschmidt, and
  Singh}]{2018executionguided}
Chenglong Wang, Po{-}Sen Huang, Alex Polozov, Marc Brockschmidt, and Rishabh
  Singh. 2018.
\newblock Execution-guided neural program decoding.
\newblock In \emph{ICML workshop on Neural Abstract Machines and Program
  Induction v2 (NAMPI)}.

\bibitem[{Warren and Pereira(1982)}]{warren1982efficient}
David~HD Warren and Fernando~CN Pereira. 1982.
\newblock An efficient easily adaptable system for interpreting natural
  language queries.
\newblock \emph{Computational Linguistics}, 8(3-4):110--122.

\bibitem[{Wong and Mooney(2007)}]{wong07}
Yuk~Wah Wong and Raymond~J. Mooney. 2007.
\newblock Learning synchronous grammars for semantic parsing with lambda
  calculus.
\newblock In \emph{Proceedings of the 45th Annual Meeting of the Association
  for Computational Linguistics (ACL-2007)}, Prague, Czech Republic.

\bibitem[{Xu et~al.(2017)Xu, Liu, and Song}]{Xu2017}
Xiaojun Xu, Chang Liu, and Dawn Song. 2017.
\newblock Sqlnet: Generating structured queries from natural language without
  reinforcement learning.
\newblock \emph{arXiv preprint arXiv:1711.04436}.

\bibitem[{Yaghmazadeh et~al.(2017{\natexlab{a}})Yaghmazadeh, Wang, Dillig, and
  Dillig}]{yaghmazadeh2017sqlizer}
Navid Yaghmazadeh, Yuepeng Wang, Isil Dillig, and Thomas Dillig.
  2017{\natexlab{a}}.
\newblock Sqlizer: query synthesis from natural language.
\newblock \emph{Proceedings of the ACM on Programming Languages}.

\bibitem[{Yaghmazadeh et~al.(2017{\natexlab{b}})Yaghmazadeh, Wang, Dillig, and
  Dillig}]{Yaghmazadeh17}
Navid Yaghmazadeh, Yuepeng Wang, Isil Dillig, and Thomas Dillig.
  2017{\natexlab{b}}.
\newblock Sqlizer: Query synthesis from natural language.
\newblock \emph{Proc. ACM Program. Lang.}, 1(OOPSLA):63:1--63:26.

\bibitem[{Yin et~al.(2018)Yin, Deng, Chen, Vasilescu, and Neubig}]{yin18msr}
Pengcheng Yin, Bowen Deng, Edgar Chen, Bogdan Vasilescu, and Graham Neubig.
  2018.
\newblock Learning to mine aligned code and natural language pairs from stack
  overflow.
\newblock In \emph{International Conference on Mining Software Repositories
  (MSR)}.

\bibitem[{Yin and Neubig(2017)}]{Yin17}
Pengcheng Yin and Graham Neubig. 2017.
\newblock A syntactic neural model for general-purpose code generation.
\newblock In \emph{{ACL} {(1)}}, pages 440--450. Association for Computational
  Linguistics.

\bibitem[{Yu et~al.(2018)Yu, Li, Zhang, Zhang, and Radev}]{Yu18}
Tao Yu, Zifan Li, Zilin Zhang, Rui Zhang, and Dragomir Radev. 2018.
\newblock Typesql: Knowledge-based type-aware neural text-to-sql generation.
\newblock In \emph{Proceedings of NAACL}. Association for Computational
  Linguistics.

\bibitem[{Zelle and Mooney(1996)}]{zelle96}
John~M. Zelle and Raymond~J. Mooney. 1996.
\newblock Learning to parse database queries using inductive logic programming.
\newblock In \emph{AAAI/IAAI}, pages 1050--1055, Portland, OR. AAAI Press/MIT
  Press.

\bibitem[{Zettlemoyer and Collins(2005)}]{Zettlemoyer05}
Luke~S. Zettlemoyer and Michael Collins. 2005.
\newblock Learning to map sentences to logical form: Structured classification
  with probabilistic categorial grammars.
\newblock \emph{UAI}.

\bibitem[{Zhong et~al.(2017)Zhong, Xiong, and Socher}]{Zhong2017}
Victor Zhong, Caiming Xiong, and Richard Socher. 2017.
\newblock Seq2sql: Generating structured queries from natural language using
  reinforcement learning.
\newblock \emph{CoRR}, abs/1709.00103.

\end{thebibliography}
\bibliographystyle{acl_natbib_nourl}

\end{document}


\maketitle

\section{SQL Hardness Criteria}

To better understand the model performance on different queries, we divide SQL queries into 4 levels: easy, medium, hard, extra hard. We define the difficulty as follows.

We first define:
\begin{itemize}
  \item SQL components 1: \texttt{WHERE, GROUP BY, ORDER BY, LIMIT, JOIN, OR, LIKE, HAVING}
  \item SQL components 2: \texttt{EXCEPT, UNION, INTERSECT, NESTED}
  \item Others: number of aggregations $>$ 1, number of select columns $>$ 1, number of where conditions $>$ 1, number of group by clauses $>$ 1, number of group by clauses $>$ 1 (no consider col1-col2 math equations etc.)
\end{itemize}

Then different hardness levels are determined as follows.

\begin{itemize}
    \item Easy: if SQL key words have ZERO or exact ONE from [SQL components 1] and SQL do not satisfy any conditions in [Others] above. AND no word from [SQL components 2].

\item Medium: SQL satisfies no more than two rules in [Others], and does not have more than one word from [SQL components 1], and no word from [SQL components 2]. OR, SQL has exact 2 words from [SQL components 1] and less than 2 rules in [Others], and no word from [SQL components 2]
\item Hard: SQL satisfies more than two rules in [Others], with no more than 2 key words in [SQL components 1] and no word in [SQL components 2]. OR, SQL has 2 $<$ number key words in [SQL components 1] $<=$ 3 and satisfies no more than two rules in [Others] but no word in [SQL components 2]. OR, SQL has no more than 1 key word in [SQL components 1] and no rule in [Others], but exact one key word in [SQL components 2].
\item Extra Hard: All others left.
\end{itemize}